\documentclass{article}

\usepackage{PRIMEarxiv}

\usepackage[utf8]{inputenc} 
\usepackage[T1]{fontenc}    
\usepackage{hyperref}       
\usepackage{url}            
\usepackage{booktabs}       
\usepackage{amsfonts}       
\usepackage{nicefrac}       
\usepackage{microtype}      
\usepackage{lipsum}
\usepackage{fancyhdr}       
\usepackage{graphicx}       
\graphicspath{{media/}}     

\title{AI Ethics Issues in Real World: Evidence from AI Incident Database
}

\author{
  Mengyi Wei \\
  Technical University of Munich \\
  \texttt{mengyi.wei@tum.de} \\
   \And
  Zhixuan Zhou \\
  University of Illinois at Urbana-Champaign \\
  \texttt{zz78@illinois.edu} \\
}

\begin{document}
\maketitle

\begin{abstract}
With the powerful performance of Artificial Intelligence (AI) also comes prevalent ethical issues. Though governments and corporations have curated multiple AI ethics guidelines to curb unethical behavior of AI, the effect has been limited, probably due to the vagueness of the guidelines. In this paper, we take a closer look at how AI ethics issues take place in real world, in order to have a more in-depth and nuanced understanding of different ethical issues as well as their social impact. With a content analysis of AI Incident Database, which is an effort to prevent repeated real world AI failures by cataloging incidents, we identified 13 application areas which often see unethical use of AI, with intelligent service robots, language/vision models and autonomous driving taking the lead. Ethical issues appear in 8 different forms, from inappropriate use and racial discrimination, to physical safety and unfair algorithm. With this taxonomy of AI ethics issues, we aim to provide AI practitioners with a practical guideline when trying to deploy AI applications ethically.   
\end{abstract}


\section{Introduction}

The comprehensive promotion of artificial intelligence (AI) technology has become a mega trend. AI has achieved satisfactory performance in some applications, yet suffers from many ethical issues, e.g. privacy violation \cite{privacy} and fake news \cite{fake}. Given the powerful transformative force of AI, and its profound influence on various sectors of society, AI ethics has drawn intense attention. More and more governments, corporations, and organizations have issued relevant guidelines to regulate AI technology and reduce ethical risks \cite{hagendorff2020ethics}. European Commission has appointed the High-Level Expert Group on Artificial Intelligence to produce reports and guidance documents on AI \cite{smuha2019eu}. In 2020, companies such as IBM and Microsoft also publicly released AI guidelines and principles \cite{ibm}. Declarations and principles have also been issued by professional associations and non-profit organizations such as the Association of Computing Machinery (ACM) \cite{meta} and UNI Global Union \cite{UNI}. 

Nevertheless, in practice, the ethical guidelines of AI are too theoretical and disjointed from practical problems, which makes it difficult to achieve the expected governance effects. For example, most AI ethics guidelines concerning transparency, justice and fairness, responsibility, and privacy, are too vague for AI practitioners to identify problems in practical applications and solve them \cite{real}. More often, system developers are unaware of the problems that will arise after the system is deployed in the real world, leading to repetitive ethical risks that are never properly addressed.

To address the above mentioned
issues with general AI guidelines, we seek to build a taxonomy of AI ethics issues in real world, including the applications areas where AI ethics issues stick out, as well as the dimensions of AI ethics issues. We approach this by conducting a comprehensive content analysis on AI Incident Database which is a catalogue of repetitive AI failures in real world \cite{database}.  

In the end, 150 AI ethics incidents were analyzed. A sharp increase of AI ethics incidents was seen after 2010, with most of them happening in the US, China, and the UK. The application areas which see most unethical behavior of AI include intelligent service robots (N=31), language/vision models (N=27), and autonomous driving (N=17), followed by intelligent recommendation (N=14), identity authentication (N=14), and AI supervision (N=14). 
AI ethics issues come in 8 different ways, including inappropriate use (bad performance), racial discrimination, physical safety, unfair algorithm (evaluation), gender discrimination, privacy, unethical use (illegal use), and mental health. To illustrate how this taxonomy can be used to guide the deployment of AI systems, we showcase the specific AI ethics issues of four different application areas.

The contribution of this work is two-fold. First, we build the first taxonomy of AI ethics issues in real world to our knowledge. Second, we provide AI practitioners with a practical handbook to guide ethical design/deployment of AI systems.

In the following sections, we will first discuss prior literature in AI ethics issues and AI ethics guidelines. Then we elaborate on our method and results. We conclude by relating our taxonomy to existing AI ethics guidelines, and reflecting on how repetitive AI failures can be mitigated. 

\section{Related Work}
AI has been applied in all areas of life \cite{recommendation:movie,recommendation:shopping,evacuation}, achieving satisfactory performance, yet suffers from numerous ethical issues, ranging from racial/gender discrimination to physical safety. Here we summarize the state-of-the-art research in AI ethics issues as well as AI ethics guidelines.

\subsection{AI Ethics Issues}
AI ethics issues manifest in various forms. While there has been no comprehensive taxonomy of unethical use of AI, here we only reflect on some outstanding AI failures which have been frequently reported and discussed.

Security and privacy are probably the most prominent issues arising from AI \cite{sp}. Vision and language models have long been known to be susceptible to adversarial attacks \cite{adversarial}, which is a malicious attempt trying to perturb data points to make them misclassified by the classifier. Though researchers worked hard to propose defenses (e.g., \cite{defense:1}, new attacks constantly emerge \cite{attack}, creating an arms force between attackers and defenders.

While user data are collected in AI-based systems for either a more precise recommendation \cite{recommender}, or a more personalized healthcare \cite{healthcare}, privacy leakage can be expected. People hold different views and perceptions toward AI privacy. For example, it was found that US people express more concerns about AI privacy, focusing more on privacy disclosure by AI applications; in contrast, Chinese people are more optimistic about AI's role in promoting privacy protection \cite{privacy:view}.

In recent year, language and vision models have always been reported to contain or amplify gender and racial bias. For example, the performance of AI-driven human facial applications is often biased toward the majority demographic group due to the data imbalance issue \cite{facial}. Nadeem et al. found that contributing factors of gender bias in AI include lack of diversity in both data and developers, programmer bias, and the existing gender bias in society which can be  amplified through AI \cite{factor}.

Despite the expanding applications of AI, and the increasing number of AI failures, there has not been a comprehensive taxonomy of AI ethics issues in real world, which is a missed opportunity to help prevent these repeated failures in the deployment of new AI systems. We bridge this research gap with the current qualitative content analysis.

\subsection{AI Ethics Guidelines}
There have long been moral concerns regarding AI systems, especially those closely knitted with people's daily life. Awad et al. deployed the Moral Machine, an online experimental platform designed to explore the moral dilemmas faced by autonomous vehicles \cite{moral}. Given the increasing number of AI failures, governments and companies have made an effort to issue AI ethics guidelines, seeking to guide the ethical design, deployment, and use of AI \cite{hagendorff2020ethics, meta, UNI}. 

Jobin et al. analysed the current corpus of principles and guidelines on ethical AI, and revealed a global convergence emerging around five ethical principles (transparency, justice and fairness, non-maleficence, responsibility, and privacy), as well as a substantive divergence regarding how these principles are interpreted, why they are deemed important, what issue, domain or actors they pertain to, and how they should be implemented \cite{meta}. 

The number of general AI ethics guidelines is increasing, and there is a large degree of convergence regarding the principles upon which these guidance documents are based. However, it is not always clear how these principles should be translated into practice. Ryan et al. tried to clarify which ethical principles could guide the development or use of AI systems, yet thought the guidelines were unlikely to have much practical effect \cite{ryan2020artificial}. In response to the limitation of previous AI ethics guidelines, we provide a more practical and nuanced understanding of how AI ethics issues happen in real world.

\section{Method}

\begin{center}
    \begin{table*}[]
        \centering
        \caption{Krippedorff's alpha for each variable. The upper part corresponds to application areas, and the lower part corresponds to AI ethics issues.}
        \begin{tabular}{c|c}
        \hline
      Content Category  & Krippendorff's Alpha \\
       \hline
      AI supervision & 0.79 \\
AI recruitment & 0.44 \\
Identity Authentication & 1 \\
Language/vision model & 0.98 \\
Intelligent recommendation & 0.96 \\
Autonomous Driving & 1 \\
Intelligent Service Robots & 1 \\
Smart Healthcare & 1 \\
AI Education & 1 \\
Predicitive policing & 1 \\
Smart Home & 1 \\
AI Game & 1 \\
Smart Finance & 1 \\
 \hline
Privacy & 1 \\
Inappropriate Use(Bad Performance) & 0.90 \\
Unethical Use(illeagal Use) & 0.97 \\
Racial Discrimination & 1 \\
Gender Discrimination & 0.98 \\
Unfair Algorithm (Evaluation) & 0.94 \\
Mental Health & 0.86 \\
Physical Safety & 1 \\
 \hline
Average & 0.94 \\
       \hline
        \end{tabular}
        \label{krippendorff}
    \end{table*}
\end{center}

\subsection{Data Collection}
We collected AI ethics incidents mainly from the AI Incident Database \cite{database}. 150 AI ethics incidents with detailed information were chosen. We identified four descriptive attributes based on a content analysis: time, geographic locations, application areas, and AI ethics issues. These four attributes cover the critical information of each AI ethics incident. 

\textbf{Time} refers to when the AI ethics incident occurred. The changing trends over time could help people understand how AI ethics issues evolve and gain/lose public attention. 

\textbf{Geographic locations} provide an overview of the global distribution of AI ethics incidents, showing the relationship between AI ethics incidents and the countries' level of development of AI technology.

\textbf{Application areas} summarize the scope of AI technology, which gives a glimpse of which areas of AI are most prone to ethical issues, and encourages more measures to solve ethical issues in these fields.

\textbf{Taxonomy of AI ethics issues} can be inferred from AI ethical incidents, which comprehensively show the unethical behavior of AI technology and the consequences for people.

\subsection{Content Analysis}
Since there was relatively little relevant research, especially regarding AI ethics issues in real world, we used a conventional approach to content analysis \cite{method:1}. We did not have preconceived categories of AI ethics issues, but instead let them emerge during the analysis.

Two authors independently coded the data, and ensured the reliability by calculating Krippendorff's alpha \cite{method:2}. The research team regularly met and discussed to refine the coding. 

In the end, the two coders reached a high agreement, with Krippendorff's Alpha larger than 0.8 on most variables, and averaging 0.94 on all variables.
The rare divergence of the two coders is on the categories of AI supervision (alpha=0.79) and AI recruitment (alpha=0.43). While one coder thought AI-decided dismissals of employees belonged to the AI recruitment category, the other coder believed they were more suitable in the AI supervision category. The subsequent results are based on the first author's coding. The agreement of the identified thirteen application areas and eight AI ethics issues is summarized in Table~\ref{krippendorff}. 

\section{Results}

Based on the content analysis, we will report on four main descriptive attributes, namely, time, geographic locations, application areas, and AI ethics issues. 
Time and location are useful attributes for understanding temporal evolution and geographical distribution of AI failures, which are rarely examined in existing research. Then we present the taxonomy of application areas and AI ethics issues. In addition, we show how AI failures in certain application areas have specific social impact, which may serve as examples to guide AI practitioners to avoid repetitive ethical issues in their work.

\subsection{Temporal Evolution of AI Ethics Incidents} 
\begin{figure}
    \centering
    \includegraphics[width=0.5\textwidth,height=0.3\textwidth]{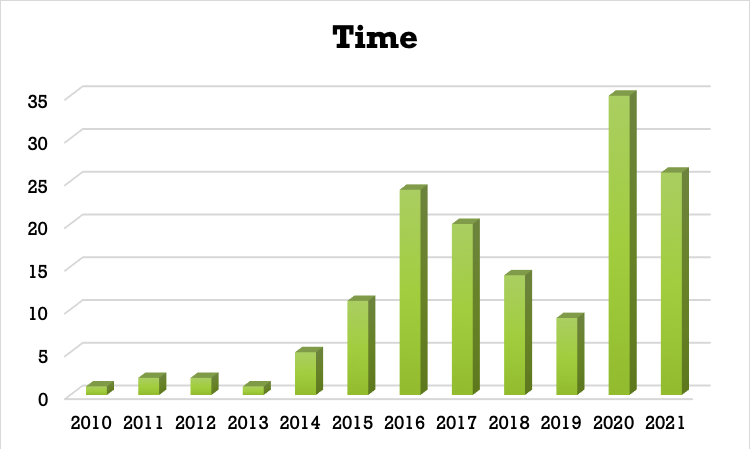}
    \caption{Temporal evolution of AI ethics incidents.}
    \label{temporal}
\end{figure}

The 150 AI ethics incidents ranged from 2010 to 2021. From Figure~\ref{temporal}, we can see that the number of AI ethics incidents gradually increased from 2010 to 2016, and reached a peak in 2016. The trend may be related to the fast development of AI during that time period: Google's AlphaGo beated world Go champion Lee Sedol; Microsoft's AI device outperformed humans in language understanding; AI made significant breakthroughs in the medical field; Tesla's self-driving vehicle sent patients to hospitals. After 2016, the number of AI ethics incidents declined until 2019, and then increased again, reaching a maximum in 2020. The trend is consistent with the heated discussion of AI ethics issues in society in recent years, which reflects people's sensitivity to the risk of AI ethics issues \cite{issue}.

\subsection{Geographic Distribution of AI Ethics Incidents} 
\begin{figure}
    \centering
    \includegraphics[width=0.5\textwidth,height=0.3\textwidth]{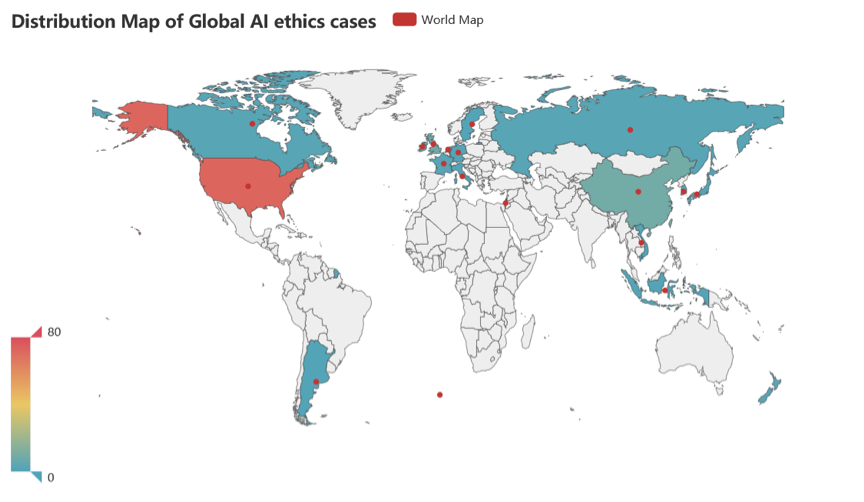}
    \caption{Geographic distribution of AI ethics incidents.}
    \label{geo}
\end{figure}

Regarding geographic distribution, we find that the most AI ethics incidents happen in the United States, China, and the United Kingdom, which account for 89 of the 150 ethical incidents (Figure ~\ref{geo}). These countries are also where most AI companies are located, such as Google (US), Tesla (US) and Baidu (China).

One of the geographic location categories is global, which means some ethical incidents occur across the world instead of being confined to a certain country. For example, Incident 14 reports gender bias embedded in the most common NLP techniques, which are used by people all over the world. The total number of global incidents is 40, demonstrating the universal nature of AI ethics failures, which deserve due attention from all countries and companies.

\subsection{Application Areas of AI Ethics Incidents} 
\begin{figure}
    \centering
    \includegraphics[width=0.5\textwidth,height=0.25\textwidth]{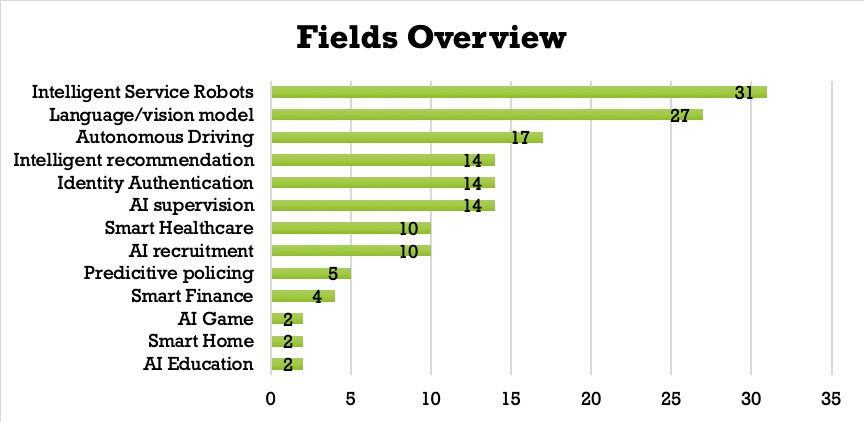}
    \caption{Application areas of AI ethics incidents.}
    \label{area}
\end{figure}

Our content analysis yielded thirteen application areas of AI which have seen ethical issues, which are, by frequency of the number of incidents: intelligent service robots, language/vision model, autonomous driving, intelligent recommendation, identity authentication, AI supervision, smart healthcare, AI recruitment, predictive policing, smart finance, AI game, smart home, and AI education (see Figure~\ref{area}). 

\textbf{Intelligence service robots} refer to a large range of robots, from manufacturing robots (Incident 5, Incident 63, Incident 64, Incident 114) to chatbots (Incident 9, Incident 56, Incident 141). The most AI ethics issues are associated with this application area (N=31, 20.4\%). Service robots pose a threat to human physical safety, causing injury or even death of workers whom they are supposed to assist. Inappropriate/biased speech of chatbots may harm the mental health of their owners.

With the prevalence of \textbf{language/vision models} also come numerous ethical failures (N=27, 17.8\%). For example, word embedding is a building block for many NLP applications, and is known to contain gender bias (Incident 14). Facebook apologized after its vision model put the `primate' label on videos of black men (Incident 134).  

The third-ranked application area is \textbf{autonomous driving} (N=17, 11.2\%). The incidents mainly report on AI ethics issues caused by the autonomous driving technology itself instead of human factors. Traffic accidents are most often caused by self-driving cars developed by Tesla (Incident 90), Uber (Incident 11), and Apple (Incident 66). 

\textbf{Intelligent recommendation} is closely related to people's daily life (N=14, 9.2\%), which has been broadly applied in online shopping \cite{recommendation:shopping}, movie recommendation \cite{recommendation:movie}, etc. Online shopping sees the most unethical use of AI. In Ctrip, a Chinese application for booking flights and hotels, users are recommended the same products with different prices given their different portraits (Incident 2). Amazon assigned lower sales rankings for books containing gay themes (Incident 17).

\textbf{Identity authentication} refers to using face recognition technologies to confirm people's identity (N=14, 9.2\%). The facial recognition system of iPhone was found susceptible to manipulation, which could be bypassed with 3D generated faces, or faces of twins (Incident 28, Incident 31). Facial recognition may also contain racial bias: a robot passport checker in New Zealand rejected an Asian man's photo for having his eyes `closed' (Incident 46). In general, unethical behaviors such as racism and sexism are often found during the authentication process (Incident 70, Incident 133, Incident 138).

\textbf{AI supervision} is used by companies to oversee, evaluate, and even monitor their employees (N=14, 9.2\%). Starbucks (Incident 3) and Amazon (Incident 91, Incident 123, Incident 131) were found using AI technology to monitor their employees' behavior, and generate excessive punishment.

Other application areas, \textbf{smart healthcare} (N=10, 6.6\%), \textbf{AI recruitment} (N=10, 6,6\%), \textbf{predictive policing} (N=5, 3.3\%), \textbf{smart finance} (N=4, 2.6\%), \textbf{AI game} (N=2, 1.3\% ), \textbf{smart home} (N=2, 1.3\%), and \textbf{AI education} (N=2, 1.3\%), also contain more or less AI ethics issues. Though the occurrences of AI ethics incidents in these application areas are relatively rare, the prevalence of AI ethics issues is clear.

\subsection{Taxonomy of AI Ethics Issues} 
\begin{figure}
    \centering
    \includegraphics[width=0.5\textwidth,height=0.27\textwidth]{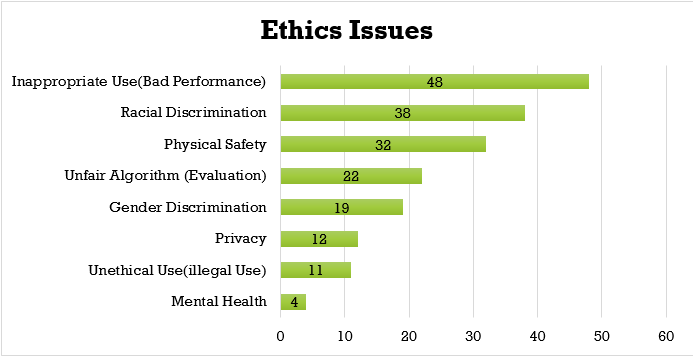}
    \caption{Taxonomy of AI ethics issues.}
    \label{consequence}
\end{figure}

Eight categories of AI ethics issues have emerged from our content analysis. These are, by frequency of occurrence: inappropriate use (bad performance), racial discrimination, physical safety, unfair algorithm (evaluation), gender discrimination, privacy, unethical (illegal use), and mental health (Figure~\ref{consequence}). It is worth noticing that one AI application may have multiple AI ethics issues. For example, a chatbot can contain gender discrimination and cause privacy leakage at the same time, showing the complicated nature of AI ethics.

Among the 150 incidents, \textbf{inappropriate use (bad performance)} is the most common issue (N=48, 25.8\%). AI technologies or algorithms often do not achieve expected performance, and even cause serious consequences. For example, some enterprises and organizations hoped to adopt AI technology to improve work efficiency (Incident 103) or provide more convenient services (Incident 63), but it turned out to be inefficient and counterproductive, for two reasons. Firstly, AI technology may not achieve the performance excepted by developers. For example, Elon Musk admitted that Tesla's production rate had been reduced due to the restriction of robots (Incident 30). Secondly, users may not have the knowledge to harness and utilize AI properly, thus fail to improve their work efficiency and performance. 

Our statistics show the prevalence of \textbf{racial discrimination} (N=38, 20.4\%), including bias, stereotyping, and imbalance against any individual based on their skin color, racial, or ethnic origin. In the scenario of predictive policing, AI technology may lean toward identifying black teenagers as criminals (Incident 139). Language/vision models deployed by tech giants such as Google, Amazon, and Facebook also make ethnic minorities unfairly treated or harmed (Incident 70, Incident 115, Incident 133, Incident 134).

The \textbf{physical safety} category focuses on risks to people's health and safety caused by AI. This issue ranks third out of eight ethics issues (N=32, 17.2\%), showing that AI technology may greatly threaten the public's physical safety if not properly designed or deployed. For example, traffic accidents are repeatedly caused by autonomous driving cars (Incident 27, Incident 90, Incident 142), and robots have been reported to hurt their human co-workers in factories (Incident 5, Incident 122).

\textbf{Unfair algorithm (evaluation)} happens when AI is used to evaluate, assess, or score people automatically (N=22, 11.8\%). Some teachers complained that their schools' AI-based evaluation algorithms unfairly gave them an `unsatisfactory' rating, which was not transparent and interpretable (Incident 12). Using AI to predict crimes has also been reported to be inaccurate, even worse than untrained human evaluators, which may cause injustice (Incident 39).

\textbf{Gender discrimination} is similar to racial discrimination, which refers to prejudice or discrimination based on one's sex or gender (N=19, 10.2\%). This issue is mainly manifested in the stereotypes embedded in algorithms. For example, in word embedding, men are associated with computer programmers, and women are associated with homemakers (Incident 14). AI algorithms may also give female candidates a lower pass rate when screening resumes in the recruitment process (Incident 20, Incident 36, Incident 45). There is also discrimination against homosexual populations (Incident 17).

The \textbf{privacy} issue is often related to companies' or organizations' desire to use personal information of users to make high profits or achieve their own goals, violating individuals' data protection rights (N=12, 6.5\%). For instance, the University of Illinois developed a remote testing software used during exams to monitor students, resulting in a privacy violation (Incident 110).

\textbf{Unethical (illegal) use of AI} is often conducted by adversaries to satisfy their needs (N=11, 5.9\%). For example, researchers used AI tools to create fake Obama videos (Incident 38). Such Deepfake techniques have long been utilized to bypass identity authentication systems, among many other malicious use cases \cite{deepfake}. Programmers created a Decentralized Autonomous Organization (DAO) in the Ethereum blockchain to steal 3.7m Ether valued at \$70m (Incident 48).

\textbf{Mental health} refers to the  emotional, psychological, and social well-being harm caused by AI technology through influencing people's cognition, perception, and behavior (N=4, 2.2\%). For example, the products recommended by Amazon's algorithm have persuade users to commit suicide (Incident 92), and a GPT-3 bot on Reddit has a strong learning ability to manipulate and deceive users of the social networking site (Incident 127).

\subsection{AI Ethics Issues in Specific Application Areas}
To understand how AI ethics issues occur in the real world, we further analyzed the proportion of AI ethics issues in each application area.

It can be seen that the most common ethical issue in intelligent service robots is inappropriate use (N=19, 48.7\%). While people expect service robots to provide better life quality or improve their work efficiency, in nearly half of the incidents, they do not achieve the expected results. Physical safety is the second most common issue in this field (N=8, 20.5\%), showing the high risk posed by service robots on users' physical safety.  





The top three ethics issues in language/vision models have been inappropriate use (N=10, 31.3\%), racial discrimination (N=10, 31.3\%), and gender discrimination (N=7, 21.9\%). Current language models have certain limitations and cannot achieve ideal results. Meanwhile, the algorithms are prone to racism or sexism. For example, Google cloud's Natural Language API provided racist, homophobic, and anti-Semitic sentiment analyses (Incident 16).   

Ethical issues in autonomous driving only involve physical safety (N=14, 77.8\%) and inappropriate use (N=4, 22.2\%), which is related to the characteristics of the field of autonomous driving. Once ethical issues arise in this field, most of them will threaten people's physical safety. Moreover, there is a class of incidents where unsatisfactory performance is caused by substandard technologies in autonomous driving, such as recognizing red words as traffic lights (Incident 90).

The most common ethical issues in the field of intelligent recommendation are racial discrimination (N=5, 26.3\%), inappropriate use (bad performance) (N=3, 15.8\%), and unethical use (illegal use) (N=3, 15.8\%). 
For example, in Incident 82, Google Images showed anti-Semitic images when a user searched ``Jewish baby stroller'', since anti-Semitic online groups have tagged these anti-Semitic images with ``Jewish baby stroller.'' The data points were unfortunately learned by Google's AI algorithm. Additionally, recommendation algorithms adopted by some enterprises will misidentify products, which are difficult to have desired effect. Facebook routinely misidentifies adaptive fashion products and blocks them from their platforms (Incident 106). The incidents related to unethical use are often misuse and even illegal use of AI technology. For example, algorithms may  recommend inappropriate content to children (on YouTube) (Incident 4) or products that make users suicidal (Incident 92). 



\section{Discussion}

\subsection{Recap of Findings}
We describe AI ethics incidents from four attributes: time, geographic locations, application areas, and AI ethics issues. From the time perspective, we analyze the trend of the number of AI ethics incidents over time and explore the reasons for their increased attention. Regarding geographical distribution, the USA, China, and the UK are the countries with the most AI ethics incidents, which are also related to the degree of AI technology development in these countries. 

We divide the application areas into 13 dimensions. Intelligent service robots, language/vision models, and autonomous driving are the areas with the most AI ethics incidents. 

Finally, we use eight categories to differentiate AI ethics issues. Among them, the most prominent problem is inappropriate use (bad performance), which explains the significant repercussions caused by the current AI technology, primarily because of the limitations of the technology itself, which cannot achieve the convenience that people expect. The issues of racial discrimination and physical safety also need to be taken seriously. Racial discrimination is relatively hidden in the algorithm, making it easy for vulnerable groups to be mistreated without even knowing. Physical safety is closely related to the (lack of) protection of human beings. If we ignore this aspect, people will be attacked by AI especially service/manufacturing robots.

\subsection{AI Ethics Incidents vs.
Guidelines}
By matching AI ethics issues in real world to AI guidelines, we find that there is consistency between them. Jobin et al. presented an overview analysis of AI guidelines. While they compiled 84 AI guideline documents, we selected the three principles with the highest frequency for a comparison \cite{meta}.

The number one ethical principle identified in \cite{meta} is transparency.
Through analysis, we find that many incidents of AI technology failure are caused by a lack of transparency. If people, even AI developers, cannot explain the exact mechanisms behind the black-box algorithms, their performance and consequence after deployment cannot be predicted and guaranteed, which leads to numerous incidents of either poor performance or malicious exploitation.

The second most frequent ethical principle occurring in AI ethics guidelines has been justice and fairness, which relates to equity, non-bias, and non-discrimination \cite{meta}. Similarly, racial and gender discrimination are found common in our analysis. According to our statistics, the area where racial discrimination occurs the most is language/vision models (N=10), which should be paid special attention to during the design, implementation and deployment process. 

The third most frequent ethical principle in previous AI guidelines is non-maleficence, relating to security, safety, harm, and protection. In our analysis, the third most common ethical issue is physical safety, echoing with the ethical guidelines. While the academia and industry have paid much attention to mitigating such ethical issues as gender and racial bias \cite{gender:bias,racial:bias}, there has been relatively rare effort in making AI algorithms physically safe to human.

Though existing ethical guidelines match ethical issues in the real world, the ethical rules are too theoretical and vague, and in most situations, people know the rules but do not know how to implement them or what the consequences will be if they do not follow the guidelines. Through analyzing AI ethics incidents in the real world, our research gives AI developers a relatively practical and concrete understanding of the severe consequences of violating the guidelines. In addition, our analysis provides a valuable perspective to guideline makers, who can formulate more operable guidelines by analyzing real incidents corresponding to the rules.

\subsection{Limitations and Outlook}
There are three main limitations of this work. Firstly, the analysis is mostly based on an existing AI incident database \cite{database}, the size and variety of which is limited. As a follow-up work, one may want to collect more AI ethics incidents in real world, for example, in news and social media, for a more thorough analysis. Secondly, we only manually analyze the AI incident database, without applying NLP models to analyze topics and sentiments in related news articles collected in the database, which is a missed opportunity. Thirdly, though the current research matches AI ethics issues with AI ethics guidelines, showing the consistency between them, and tries to solve the ambiguity of the principles, considerably more work is needed to refine the theoretical and operable parts of the guidelines. With the analysis of AI ethics incidents in real world, we make the first step toward assisting principle makers in formulating more practical guidelines.

\section{Conclusion}
In this paper, we picture the landscape of real-world AI ethics incidents with an exploratory content analysis. Intelligent service robots, language/vision models, and autonomous driving are the three application areas where AI failures occur most frequently. AI ethics issues span all dimensions of people's life, including racial/gender discrimination, physical safety, privacy leakage, etc. By closely inspecting AI ethics issues associated with the top four application areas, we provide an example for AI practitioners to mitigate AI ethics issues in their work. We also relate AI ethics incidents to AI ethics guidelines, and provide a perspective for guideline makers to formulate more operable guidelines by analyzing real-world incidents corresponding to the rules.


\bibliographystyle{unsrt}  
\bibliography{references}  

\end{document}